%% file: acl_latex.tex
\pdfoutput=1

\documentclass[11pt]{article}

\usepackage[preprint]{acl}

\usepackage{times}
\usepackage{latexsym}

\usepackage[T1]{fontenc}

\usepackage[utf8]{inputenc}

\usepackage{microtype}

\usepackage{inconsolata}

\usepackage{graphicx}
\usepackage{subcaption}
\usepackage{wrapfig}

\usepackage{amsmath}
\usepackage{amstext}
\usepackage{amsfonts}
\usepackage{bm}
\usepackage{bbm}

\usepackage{booktabs}
\usepackage{multirow}
\usepackage{array}
\usepackage{caption}
\usepackage{color}
\usepackage{xcolor}
\usepackage{colortbl}
\usepackage{tablefootnote}
\usepackage{adjustbox}

\usepackage{caption}
\usepackage{algpseudocode}
\usepackage[ruled,noend]{algorithm2e}

\usepackage{xspace}
\usepackage{xcolor}

\def\ours{CLaSp\xspace}
\def\vm{verify model\xspace}
\def\dm{draft model\xspace}

\definecolor{myblue}{RGB}{210, 220, 250}
\definecolor{myorange}{RGB}{237,125,49}
\definecolor{myred}{RGB}{148,17,0}
\definecolor{darkgreen}{rgb}{0.0, 0.5, 0.0}
\definecolor{darkred}{rgb}{0.55, 0.0, 0.0}

\newcommand{\orange}{\cellcolor{myorange!10}}
\newcommand{\ccmt}[1]{\textcolor[RGB]{64,101,149}{#1}}

\title{CLaSp: In-Context Layer Skip for Self-Speculative Decoding}

\author{%
  \textbf{Longze Chen}$^{1,2}$\thanks{Equal contribution.}\quad
  \textbf{Renke Shan}$^{2,5}$\footnotemark[1]\quad
  \textbf{Huiming Wang}$^{3}$\footnotemark[1]\quad
  \textbf{Lu Wang}$^{5}$\quad
  \textbf{Ziqiang Liu}$^{1,2}$\\
  \textbf{Run Luo}$^{1,2}$\quad
  \textbf{Jiawei Wang}$^{2}$\quad
  \textbf{Hamid Alinejad-Rokny}$^{4}$\quad
  \textbf{Min Yang}$^{1}$\thanks{Corresponding author.}\\  
  $^{1}$ Shenzhen Institutes of Advanced Technology, Chinese Academy of Sciences\\
  $^{2}$ University of Chinese Academy of Sciences\quad
  $^{3}$ Singapore University of Technology and Design\\ 
  $^{4}$ School of Biomedical Engineering, UNSW Sydney\quad
  $^{5}$ Ritzz-AI\\
  \texttt{\{lz.chen2, min.yang\}@siat.ac.cn}
}

\begin{document}
\maketitle 
\begin{abstract}
Speculative decoding (SD) is a promising method for accelerating the decoding process of Large Language Models (LLMs). 
The efficiency of SD primarily hinges on the consistency between the \dm and the \vm. 
However, existing drafting approaches typically require additional modules to be trained, which can be challenging to implement and ensure compatibility across various LLMs.
In this paper, we propose \textbf{CLaSp}, an in-context layer-skipping strategy for self-speculative decoding. 
Unlike prior methods, \ours does not require additional drafting modules or extra training. 
Instead, it employs a plug-and-play mechanism by skipping intermediate layers of the \vm to construct a compressed \dm.
Specifically, we develop a dynamic programming algorithm that optimizes the layer-skipping process by leveraging the complete hidden states from the last verification stage as an objective.
This enables \ours to dynamically adjust its layer-skipping strategy after each verification stage, without relying on pre-optimized sets of skipped layers.
Experimental results across diverse downstream tasks demonstrate that \ours achieves a speedup of $\bm{1.3\times\sim1.7\times}$ on LLaMA3 series models without altering the original distribution of the generated text.
\end{abstract}

\begin{figure}[t]
  \centering
  \includegraphics[width=1.0\columnwidth]{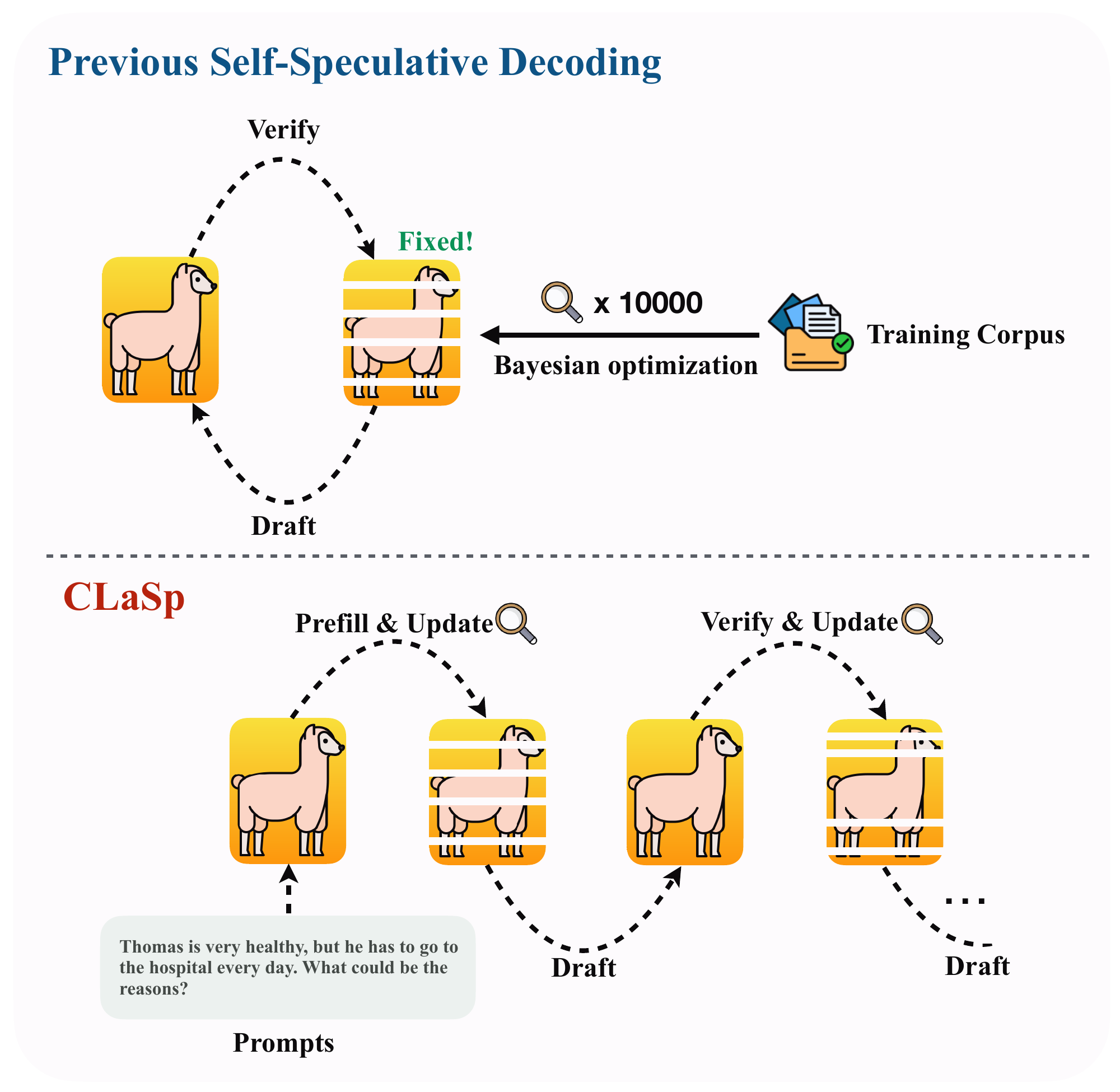}
  \caption{\textbf{Previous Self-SD method vs. \ours}. Compared to the previous Self-SD method, which requires costly Bayesian optimization on training dataset to select a {\it \textbf{fixed}} set of skipped layers, \ours employs a {\it \textbf{dynamic}} layer-skipping strategy that adjusts in real-time based on context.}
  \label{fig:overview}
\end{figure}

\section{Introduction}
Transformer-based Large Language Models (LLMs) have achieved remarkable success across a wide range of natural language processing applications~\citep{brown-2020-language, achiam-2023-gpt4}. 
Scaling the model size and extending the context window significantly enhance performance~\citep{kaplan-2020-scaling, anil-2023-palm, reid-2024-gemini}, but also leads to a rapid increase in inference latency.
This latency primarily stems from the autoregressive nature of LLMs, where model parameters must be loaded into GPU SRAM for each token generation, resulting in underutilization of computational cores during the decoding stage~\citep{patterson-2004-latency, shazeer-2019-fast, agrawal-2023-sarathi}.

Inspired by speculative execution in computer systems~\citep{burton-1985-speculative, hennessy-2017-computer}, speculative decoding (SD)~\citep{xia-2023-speculative, leviathan-2023-fast, chen-2023-accelerating} is proposed as a lossless autoregressive decoding acceleration technique.
SD accelerates autoregressive decoding by introducing an efficient \dm to pre-generate tokens, which are subsequently validated by a slower \vm in parallel. 
This technique significantly reduces the number of forward passes required by the \vm, alleviating memory-bound inefficiencies caused by frequent parameter access. 
While effective, SD relies on finding or training a suitable \dm that can closely mimic the behavior of the \vm.
This requirement is feasible for open-sourced model families, such as LLaMA series~\citep{touvron-2023-llama, touvron-2023-llama2, dubey-2024-llama3, yang-2024-qwen2}, but becomes prohibitively difficult for specialized LLMs that lack pre-existing compatible \dm counterparts.

The difficulty lies in achieving consistency between the \dm and the \vm. 
For general-purpose models, lightweight modules have been proposed as substitutes for the \dm~\citep{cai-2024-medusa, li-2024-eagle, du-2024-glide, liu-2024-kangaroo}. 
These modules are designed to avoid retraining from scratch, but they struggle to generalize across diverse tasks and contexts.
As a result, their acceptance rates drop sharply when handling unseen tasks, making them unsuitable for applications requiring robust performance across varying scenarios. 

Self-speculative decoding (Self-SD)~\citep{zhang-2024-draft} addresses the challenge of compatibility by using parts of the \vm itself as a compressed \dm, bypassing the need for additional modules or training. 
This approach creates the \dm by sparsifying intermediate layers of the \vm, effectively skipping certain computations. 
Similar to methods that require training, it also lacks robust generalization and heavily relies on an time-consuming Bayesian optimization process.
SWIFT~\citep{xia-2024-swift} extends Self-SD by dynamically optimizing skipped layers as the number of user requests increases, but its effectiveness diminishes when handling sparse or unique task data.

To bridge this gap, we propose \textbf{CLaSp}, a dynamic in-context layer-skipping method for self-speculative decoding. 
Unlike existing methods, \ours dynamically adjusts the skipped layer set at each decoding step based on the current context, eliminating the need for pre-optimization or retraining (see Figure~\ref{fig:overview}). 
Our approach leverages the observation of slowly changing embeddings across layers~\citep{liu-2023-dejavu} and employs a dynamic programming algorithm to identify the optimal skipped layers with minimal additional latency. 
By using the complete hidden states from the last verification step as ground truth, \ours predicts and adjusts the \dm's sparsity in real-time, achieving high acceptance rates while maintaining acceleration benefits.

We evaluate \ours on the LLaMA3 series models using Spec-Bench~\citep{xia-2024-unlocking}, a comprehensive benchmark for speculative decoding across diverse scenarios. 
\ours achieves $1.3 \times \sim 1.7 \times$ wallclock time speedup compared to conventional autoregressive decoding while preserving the original distribution of generated text. 
Our contributions are summarized as follows:
\begin{itemize}
\item We introduce \ours, a self-speculative decoding framework that dynamically adjusts the layer-skipping strategy based on context.
\item We propose performance optimization strategies in \ours to fully leverage GPU parallelism, making the extra latency from layer optimization almost negligible.
\item We conduct extensive experiments on Spec-Bench, showing that \ours consistently achieves $1.3\times \sim 1.7\times$ speedup without training, and provide a detailed analysis of its key hyper-parameters.
\end{itemize}

\section{Related Work}

\paragraph{Speculative Decoding.}
Speculative decoding~\citep{xia-2023-speculative, leviathan-2023-fast, chen-2023-accelerating} has been proposed as an effective strategy for lossless acceleration of LLM inference.
Some approaches aim to reduce the high cost of training from scratch by introducing lightweight modules as \dm.
Medusa~\citep{cai-2024-medusa} trains multiple decoding heads to predict the next $n$ tokens in parallel. 
EAGLE and EAGLE-2~\citep{li-2024-eagle, li-2024-eagle2} integrate a lightweight plug-in (a single transformer decoder layer) to existing LLMs.
Glimpse Draft Model~\citep{du-2024-glide} reuses the \vm's KV cache to generate candidate tokens that are more likely to be accepted by the \vm.
However, these approaches rely heavily on pre-trained modules optimized for specific contexts, making them less effective for tasks with unseen data. 

Retrieval-based methods, such as REST~\citep{he-2024-rest} and Prompt Lookup Decoding~\citep{saxena-2023-prompt}, replace the \dm by retrieving relevant drafts from a text corpus or context based on input prompts. 
While these methods reduce reliance on explicit \dm models, their performance is highly sensitive to the quality and relevance of the retrieved drafts. 

To address the challenges of designing compatible \dm, Self-SD~\citep{zhang-2024-draft} and SWIFT~\citep{xia-2024-swift} directly leverage parts of the \vm as a compressed \dm by skipping intermediate layers of the original LLM. 
Triforce~\citep{sun-2024-triforce} employs a partial KV cache as the \dm and a full KV cache as the \vm, reducing inference latency by minimizing I/O operations, especially in long-context tasks. 
Despite these innovations, static configurations for layer skipping prevent these methods from dynamically adapting to changing task requirements or contexts, limiting their efficiency and generalizability.

To further enhance speculative decoding, tree-based attention mechanisms~\citep{miao-2023-specinfer, cai-2024-medusa, chen-2024-sequoia, svirschevski-2024-specexec} extend the decoding process from generating a single candidate sequence to exploring a candidate tree. 
By providing the \vm with multiple options for validation, these methods improve the acceptance rate of speculative decoding.

\paragraph{Layer-wise Sparsity.}
Layer redundancy in LLMs has been extensively studied, with methods such as LayerDrop~\citep{angela-2020-reducing}, LayerSkip~\citep{elhoushi-2024-layerskip}, structured pruning~\citep{zhang-2020-accelerating}, SkipDecode~\citep{luciano-2023-skipdecode}, and LayerSharing~\citep{zhang-2023-crash} demonstrating that not all layers are equally important. 
These methods suggest that the importance of each layer varies depending on the task and context, and that certain layers can be skipped or removed without significantly affecting model performance. 
However, determining which layers to skip or optimize for different downstream tasks remains a substantial challenge.

Deja Vu~\citep{liu-2023-dejavu} and LISA~\citep{pan-2024-lisa} demonstrate the potential of leveraging sparsity to accelerate LLM inference, either by exploiting context sparsity or by optimizing a subset of layers during training. 
Although effective, these methods rely on lossy sparsification techniques, introducing discrepancies between the sparse and original distributions. 
Similarly, \citet{glavas-2024-dynamic} explore dynamic inference methods such as layer skipping and early exiting, which enable task-dependent acceleration but lack compatibility with speculative decoding, limiting their potential for lossless acceleration.
Thus, we aim to combine the strengths of layer-wise sparsity and speculative decoding to achieve lossless acceleration across diverse tasks.

\input{algs/clasp_dp}

\begin{figure*}[t]
  \centering
  \includegraphics[width=1.0\textwidth]{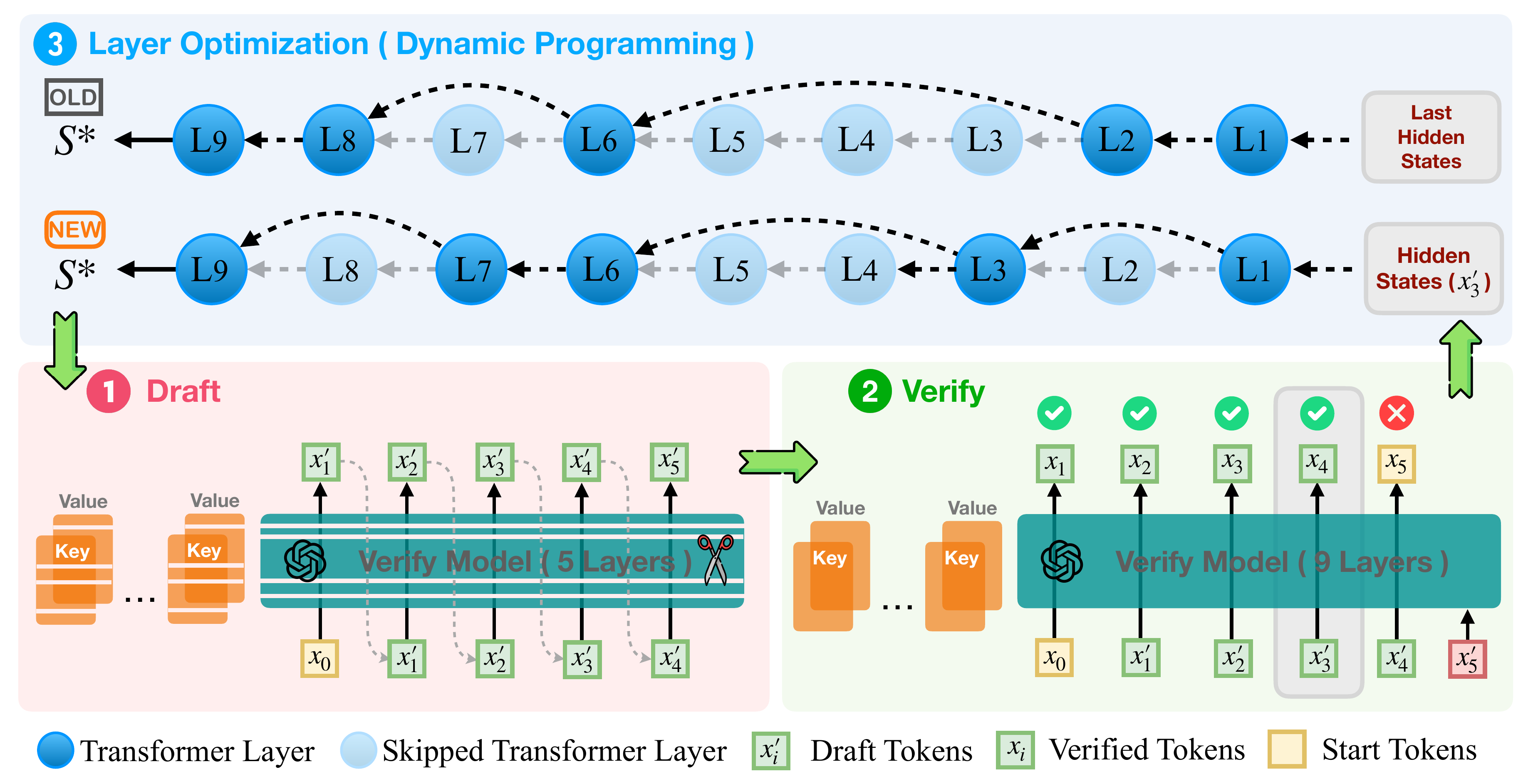}
  \caption{The overall framework of \ours consists of three stages: (1) Draft, (2) Verify, (3) Layer Optimization. After the Verify stage, \ours uses the information obtained to perform Layer Optimization, resulting in a new optimal layer skipping set $\mathcal{S}^*$. This set guides the next Draft round, repeating the entire process.}
  \label{fig:framework}
\end{figure*}

\section{CLaSp}
In this section, we first introduce the pipeline of \ours from a global perspective.
Then, we explore the main challenges~(\S\ref{sec:mainchallenge}) faced by \ours and formulate the problem of layer skipping~(\S\ref{sec:problmeform}). 
Subsequently, we provide a detailed description of the \ours algorithm~(\S\ref{sec:adp} and \S\ref{sec:amp}) and efficiency optimization strategies~(\S\ref{sec:parallel} and \S\ref{sec:lof}).

\subsection{Pipeline}
\label{sec:pipeline}
\ours can be explained as a three-stage process: 
(1) \textbf{Drafting}: The \dm autoregressively generates $K$ draft tokens from the given prompt sequence $x_1, \ldots, x_i$, denoted as $x_{i+1}, \ldots, x_{i+K}$.
(2) \textbf{Verification}: The \vm verifies the tokens generated during the drafting stage. This verification is performed in a single forward pass, where the LLM predicts the probability distribution for each draft token and evaluates whether they align with the full model's predictions. Once a draft token $x_j$ is rejected, the original LLM's prediction overwrites $x_j$, and drafting resumes from token $x_{j+1}$ in the next round.
(3) \textbf{Layer Optimization}: Using the hidden states of the last accepted token $x_{j}$ as the optimization objective, the optimal skipped layer set $\mathcal{S}^{*}$ is updated to guide the next round of drafting. As shown in Figure~\ref{fig:framework}, before each round of drafting, the \dm can be updated to better adapt to the current context.

\subsection{Main Challenges}
\label{sec:mainchallenge}
Compared to previous methods, \ours dynamically updates the skipped layer set before each drafting step, requiring solutions to two main challenges:

{\it \textbf{(1) How to determine which layers should be skipped?}}
This is the most critical issue addressed by \ours, as it directly impacts drafting quality. 
An ideal layer-skipping strategy must adapt to the most recent context, ensuring that the drafted tokens are more likely to be accepted by the \vm.

{\it \textbf{(2) How to reduce the additional latency caused by layer optimization?}}
The dynamic skipping strategy inevitably introduces computational delays due to the need for repeated searches to identify the current optimal layer subset. 
To ensure that layer optimization does not become the primary bottleneck, minimizing these additional delays is essential for maximizing the speedup benefits.

\subsection{Problem Formulation of Layer Skip}
\label{sec:problmeform}
Let $\mathcal{M}_{v}$ be the \vm and $\mathcal{M}_{d}$ be the \dm obtained by skipping certain intermediate layers from the original LLM.
$F_{\mathcal{M}_{v}}(X)$ and $F_{\mathcal{M}_{d}}(X)$ represent the output hidden states on the top of the last token of current input $X$, passing through the \vm or the \dm respectively.
Our goal is to find the optimal skipped layer set $\mathcal{S}$ that minimizes the cosine similarity between $F_{\mathcal{M}_{v}}(X)$ and $F_{\mathcal{M}_{d}}(X)$:
\begin{equation}
\begin{aligned}\label{eq:goal}
\mathcal{S}^*=\underset{\mathcal{S}}{\arg \min } \; cosine\textcolor{violet}{(}F_{\mathcal{M}_{V}}(X), F_{\mathcal{M}_{D}}(X)\textcolor{violet}{)}, \\ \text { s.t. } \mathcal{S} \in\{0,1\}^L
\end{aligned}
\end{equation}
where $L$ represents the number of transformer layers in the \vm.

\begin{figure*}[t]
  \centering
  \includegraphics[width=1.0\textwidth]{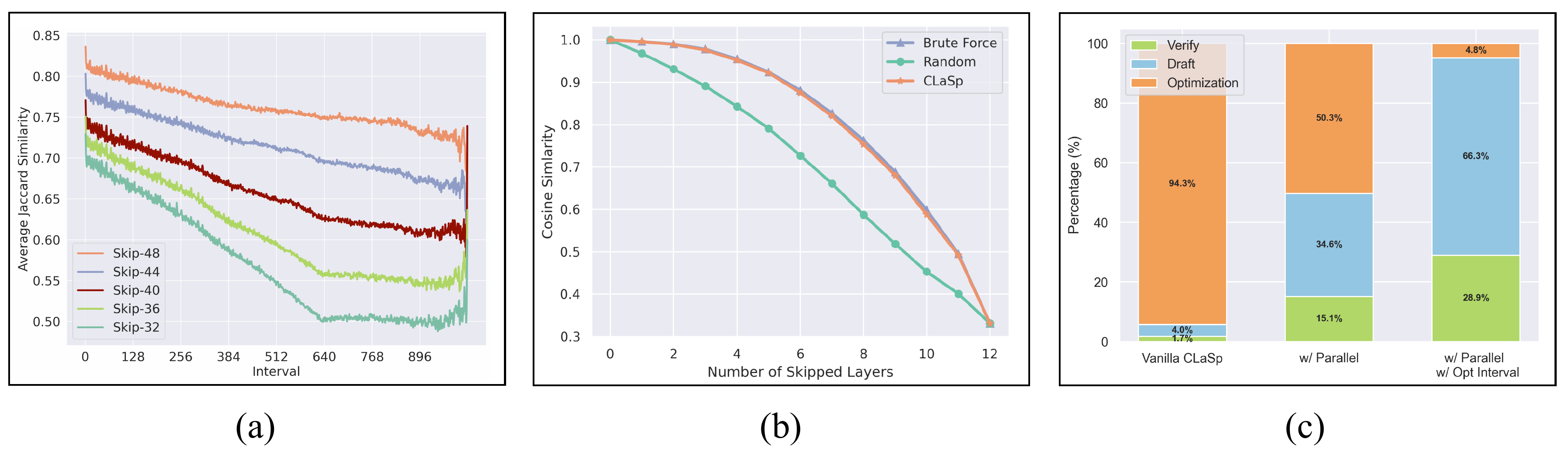}
  \caption{(a) \textbf{Sparse Persistence Observation}: Skipped layer sets selected for adjacent tokens exhibit high similarity, with this similarity gradually decreasing as the token gap increases. This observation enables layer optimization on the current token to guide subsequent drafting processes. 
  (b) \textbf{Approximate Markov Property}: Cosine similarity comparisons of hidden states obtained using Brute Force, Random, and \ours's dynamic programming configurations against the full forward pass demonstrate the approximate Markov property inherent to \ours. 
  (c) \textbf{Efficiency Optimization Strategies}: Latency breakdown per query shows that Layer Optimization introduces only 4.8\% additional delay, underscoring its negligible impact on overall latency.}
  \label{fig:proof}
\end{figure*}

\subsection{Approximate Dynamic Programming}
\label{sec:adp}
The principle behind selecting information for layer optimization is to minimize additional computational overhead by utilizing information already obtained in previous steps. 
In speculative decoding, we observed that the hidden states of the last accepted token after each verification step are not fully utilized.
To address this, we propose leveraging this feedback information to predict the \dm configuration for the next drafting stage.
Specifically, let the input tokens to a Transformer model be denoted as $X$, with an embedding layer that maps token indices to token embeddings $h_0$. The Transformer model consists of $L$ layers, where the $l$-th Transformer layer performs a transformation $f_l$, evolving the embeddings as: 
\[
h_{l+1} = f_l\textcolor{cyan}{(}h_l\textcolor{cyan}{)}
\]

Let $\mathcal{D}(i, j)$ represent the maximum cosine similarity between $h_i$ and the optimal hidden state $g\textcolor{orange}{(}i, j\textcolor{orange}{)}$ obtained by skipping $j$ layers among the first $i$ transformer layers. So we design a dynamic programming transition equation defined as:
\begin{equation}
\begin{aligned}\label{eq:dp}
\mathcal{D}(i, j) = \max\{cosine\textcolor{violet}{(}h_i, g\textcolor{orange}{(}i - 1, j - 1\textcolor{orange}{)}\textcolor{violet}{)}, \\ cosine\textcolor{violet}{(}h_i, f_{i-1}\textcolor{cyan}{(}g\textcolor{orange}{(}i - 1, j\textcolor{orange}{)}\textcolor{cyan}{)}\textcolor{violet}{)}\}
\end{aligned}
\end{equation}

where $cosine$ is used to calculate the cosine similarity between two vectors.
The \ours skip layer algorithm process is shown in Algorithm~\ref{alg:CLaSp}.

\subsection{Approximate Markov Property}
\label{sec:amp}
A crucial prerequisite for dynamic programming algorithms is the "no aftereffect" property, which ensures that current decisions and state transitions are independent of previous states.
However, when computing the optimal hidden states $g\textcolor{orange}{(}i, j\textcolor{orange}{)}$, \ours does not strictly satisfy the Markov property, making it theoretically impossible to find an exact optimal solution using Algorithm~\ref{alg:CLaSp}.

Fortunately, due to the favorable property of slowly changing embeddings across layers, we observe that \ours's approximate algorithm closely aligns with the results of a brute force search for the optimal skipped layer set.
To validate this, we fixed the first and last 10 layers of the 32-layer LLaMA3-8B model and compared the outcomes of brute force search, random layer selection, and \ours across the remaining 12 layers. 

As shown in Figure~\ref{fig:proof}b, the hidden states obtained by skipping the layers selected by \ours exhibit remarkable consistency with those from the brute force search, demonstrating a high cosine similarity with the hidden states of the original LLM. 
In contrast, the results from randomly selected layers show relatively poor alignment. 

These findings indicate that \ours approximates the Markov property effectively, allowing it to find near-optimal solutions within an acceptable error range.

\subsection{Sequence Parallel}
\label{sec:parallel}
Unlike previous methods, \ours requires multiple layer optimizations during a single inference process. 
Therefore, the optimization process must be both efficient and accurate to avoid introducing additional delays while ensuring precise drafting in subsequent decoding steps.
To address this, we employ parallelization strategies to minimize the additional delay caused by the dynamic programming process.

When \ours performs dynamic programming, the updates for $\mathcal{D}(i, j)$ and $g\textcolor{orange}{(}i, j\textcolor{orange}{)}$ are obtained through a double loop, resulting in a time complexity of $\mathcal{O}(LM)$.
Importantly, when computing the state at $(i, j)$, only the state at $(i - 1, \cdot)$ is required.
This dependency allows computations for different $j$ values with the same $i$ to be performed independently, enabling parallelization of the second loop.

To further reduce the GPU memory footprint, we avoid concatenating these states into a batch. Instead, we design a specialized mask matrix that enables parallelization of these states as a sequence. This approach reuses the same KV cache without duplicating it, significantly improving memory efficiency.

\input{tabs/main}

\subsection{Lower Optimization Frequency}
\label{sec:lof}

\ours updates the optimal skipped layer set after each verification step, using the feedback from the last accepted token. 
However, the time cost of performing this update is comparable to that of a verification step, which introduces a bottleneck for \ours's inference latency.
Fortunately, we observe a phenomenon we term \textbf{Sparse Persistence}: the skipped layer sets required by adjacent tokens tend to exhibit high similarity. 
To quantify this, we calculate the Jaccard similarity between the skipped layer sets of adjacent tokens.
As shown in Figure~\ref{fig:proof}a, the similarity remains high when the token distance is within a certain range and only decreases significantly as the distance between tokens increases.

Based on this observation, we further found that the optimal skipped layer set does not change drastically after every update.
This allowed us to adjust the update frequency by accumulating several verification steps before performing an update, rather than updating after every single verification step.
While adopting a lower update frequency slightly reduced the average acceptance rate of draft tokens, the reduction in update latency led to a substantial improvement in the overall speedup ratio. 
This trade-off highlights the efficiency benefits of leveraging Sparse Persistence to optimize the layer update process.

\begin{figure*}[t]
  \centering
  \begin{subfigure}[b]{0.325\linewidth}
      \includegraphics[width=1.0\linewidth]{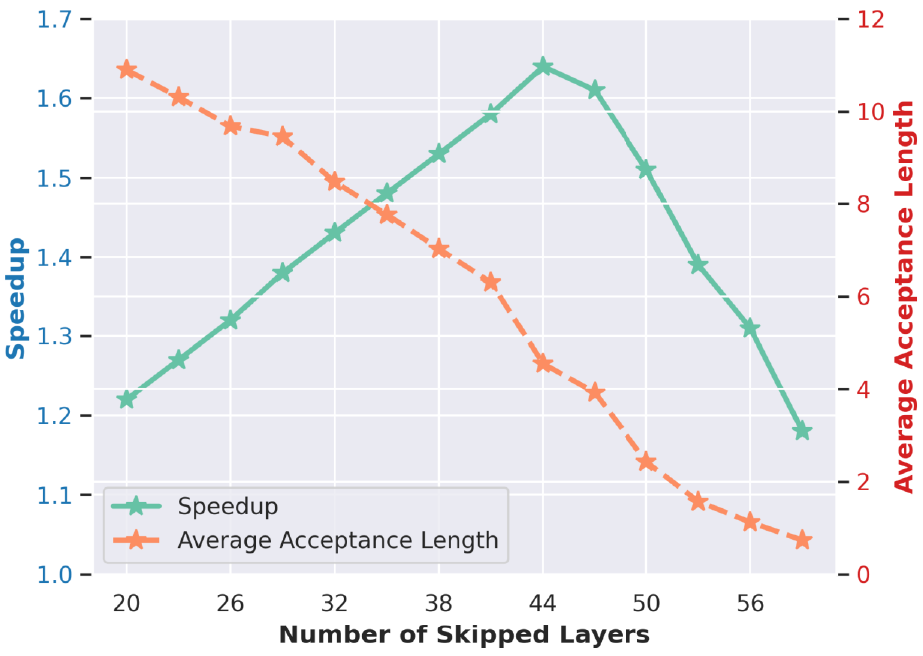}
      \caption{}
      \label{fig:layer-skip-ratio}
  \end{subfigure}
  \hfill
  \begin{subfigure}[b]{0.325\linewidth}
      \includegraphics[width=1.0\linewidth]{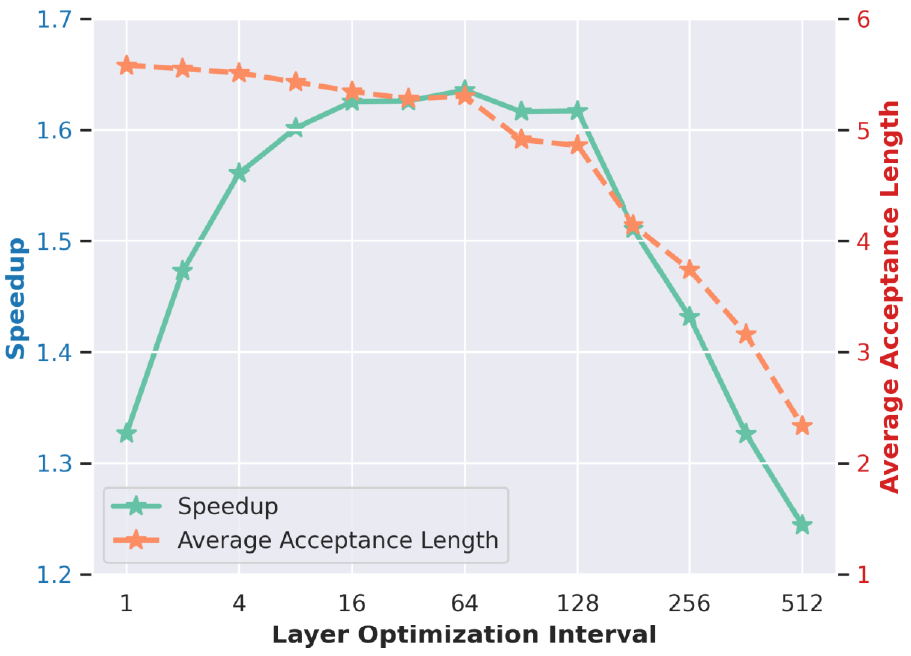}
      \caption{}
      \label{fig:layer-opt-interval}
  \end{subfigure}
  \hfill
  \begin{subfigure}[b]{0.325\linewidth}
      \includegraphics[width=1.0\linewidth]{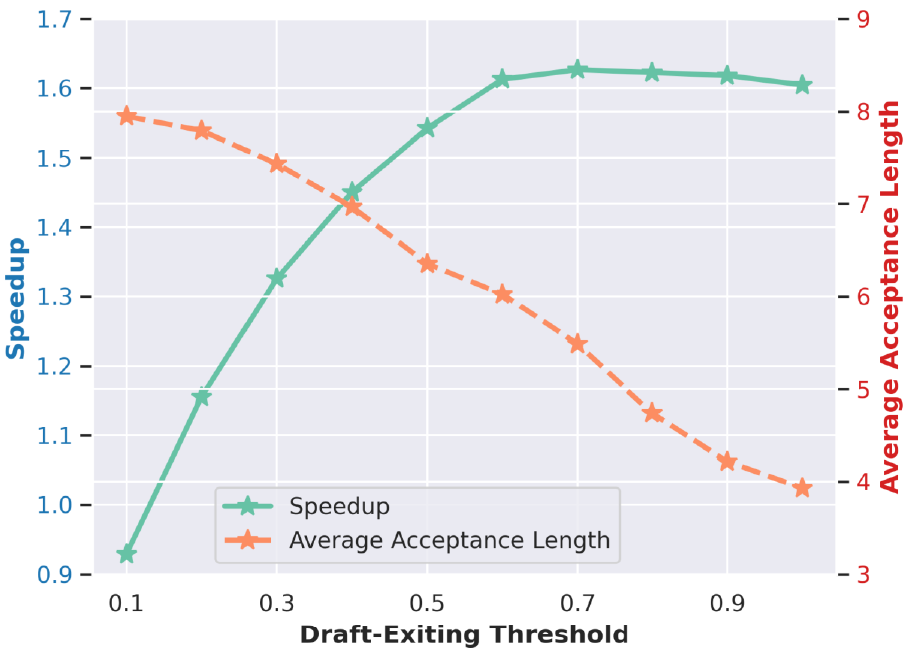}
      \caption{}
      \label{fig:draft-existing-threshold}
  \end{subfigure}
  \caption{The impact of key hyper-parameters on speedup: (a) Number of Skipped Layers; (b) Layer Optimization Interval; (c) Draft-Existing Threshold. The experiment results were all obtained using the LLaMA-3-70B model on MT-Bench.}
\end{figure*}

\section{Experiments}
This section evaluates \ours across various text generation tasks to demonstrate its efficiency and effectiveness.

\paragraph{Model and Testbed.}
We evaluate \ours using four different sizes of LLaMA models~\citep{dubey-2024-llama3}: LLaMA3-8B, LLaMA2-13B, LLaMA3-70B, and LLaMA3.1-405B. 
The models are deployed on NVIDIA A800 GPUs with 80GB of memory. 
Specifically, the 8B and 13B models are deployed on a single A800 GPU, while the 70B and 405B models utilize 2 and 8 A800 GPUs, respectively, with pipeline parallelism enabled by Accelerate~\citep{gugger-2022-accelerate}. 
All models use FP16 precision, except for LLaMA3.1-405B, which employs INT8 quantization for improved memory efficiency. 
Unless otherwise specified, the batch size is set to 1 for all models.

\paragraph{Datasets.}
We benchmark the performance of \ours on Spec-Bench~\citep{xia-2024-unlocking}, a comprehensive evaluation suite covering a wide range of datasets and tasks. 
Spec-Bench includes six subtasks: multi-turn conversation, translation, summarization, question answering, mathematical reasoning, and retrieval-augmented generation. 
Specifically, Spec-Bench consists of 80 randomly selected instances from each of MT-bench~\citep{zheng-2023-judging}, WMT14 DE-EN, CNN/Daily Mail~\citep{nallapati-2016-abstractive}, Natural Questions~\citep{kwiatkowski-2019-natural}, GSM8K~\citep{cobbe-2021-training}, and DPR~\citep{karpukhin-2020-dense}.
To control the generation length across tasks, we set the maximum sequence length to 1024 tokens, following prior experimental setups~\citep{xia-2024-unlocking}.

\paragraph{Comparison.}
For our main experiments, we use vanilla autoregressive decoding as the baseline, serving as the benchmark for speedup ratios (1.00x). 
We compare \ours against existing training-free layer skip methods, including Self-Speculative Decoding~\citep{zhang-2024-draft} and SWIFT~\citep{xia-2024-swift}. 
Other speculative decoding (SD) methods are excluded from the comparison as they require additional modules or extensive training, which limits their generalizability. 
Since the speedup ratio is hardware-dependent, all methods were evaluated on the same devices to ensure a fair comparison.

\paragraph{Performance Metrics.}
\ours is essentially still speculative sampling, which has been proven to enable lossless acceleration~\citep{leviathan-2023-fast}. 
Therefore, we focus solely on acceleration performance rather than generation quality. 
The following metrics are used for evaluation:
\textbf{Speedup Ratio}: The actual test speedup ratio relative to vanilla autoregressive decoding, providing a direct measure of acceleration. 
\textbf{Average Acceptance Length ($\tau$)}: The average number of tokens generated per drafting-verification cycle, corresponding to the number of tokens accepted from the draft. This metric is independent of hardware and runtime environment, while its limitation is that it does not reflect the overhead introduced by the \dm.

\subsection{Experimental Result}
As shown in Table~\ref{tab:main-exp}, we report the performance of \ours and prior plug-and-play methods on text generation tasks from Spec-Bench under both greedy (Temperature = 0) and non-greedy (Temperature = 1) settings. 
The experimental results reveal the following findings.

\ours demonstrates superior efficiency compared to previous methods, achieving consistent speedups of $1.3\times\sim1.7\times$ over vanilla autoregressive decoding across various models and tasks. 
Prior methods relying on Bayesian optimization exhibit lower performance, particularly when data volume is limited.

\ours consistently improves average acceptance length, acceptance rate, and speedups. This efficiency is primarily due to \ours's ability to leverage model layer sparsity effectively. By skipping 50\% to 60\% of layers during experiments, \ours maintains both a high average acceptance length and acceptance rate, contributing to superior acceleration. Generally, longer acceptance lengths lead to higher speedups. However, there are cases where speedups remain low despite long acceptance lengths, as drafting additional tokens increases time spent, reducing acceptance rates and overall speedups. 

The performance advantage of \ours is more pronounced on larger models, such as LLaMA3-70B, compared to smaller models like LLaMA2-13B and LLaMA3-8B. This suggests that \ours can better leverage the greater layer sparsity present in larger models, enhancing adaptability and efficiency. 

Overall, the robust performance of \ours across different models highlights its effectiveness as a plug-and-play solution, offering a versatile method to enhance inference speed for a range of LLMs.

\begin{figure}[t]
  \centering
  \includegraphics[width=1.0\columnwidth]{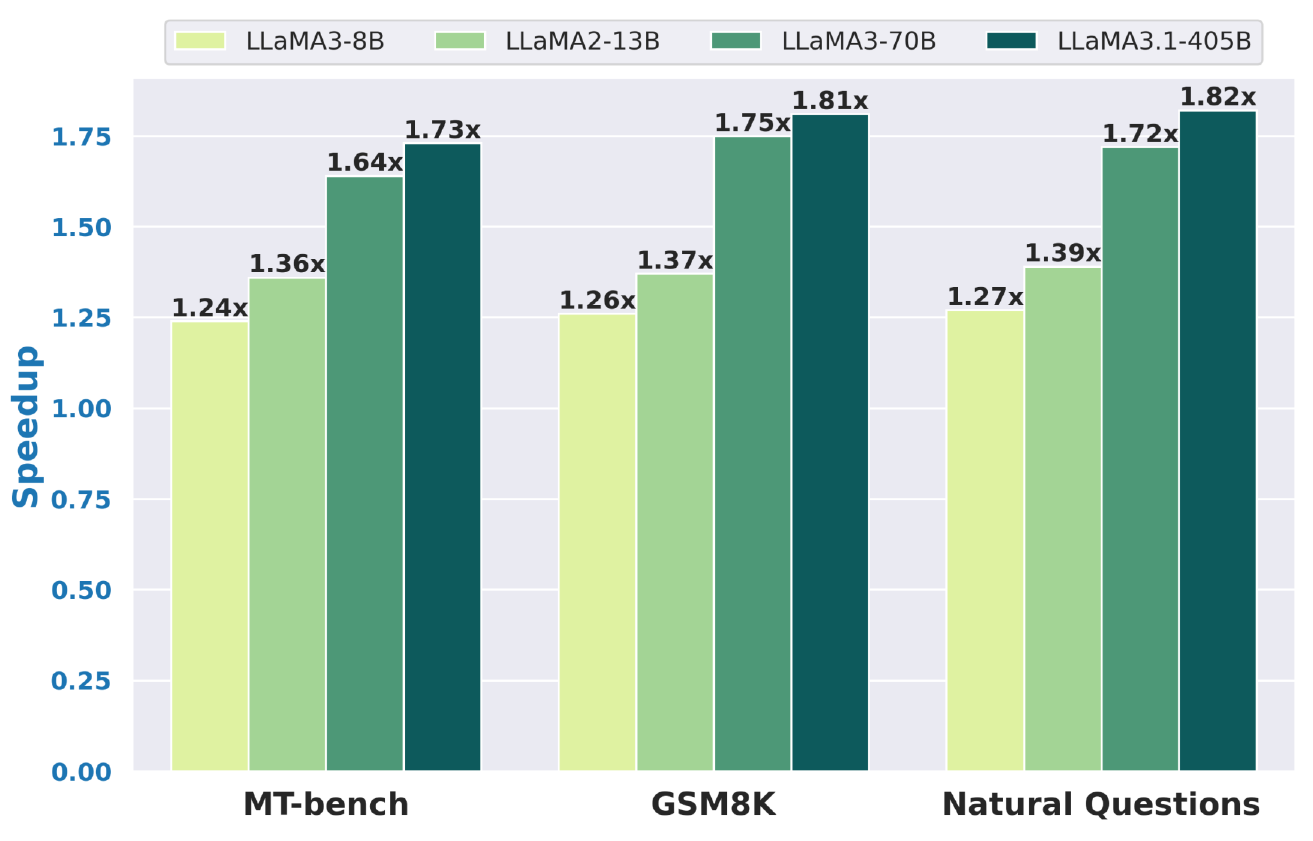}
  \caption{Model Size Scaling Laws of \ours.}
  \label{fig:different-model}
\end{figure}

\section{Analysis}
We present an extensive analysis of \ours, focusing on three key aspects: the benefits of the parallel strategy~(Section~\ref{ab:sp}), compatibility with different LLMs~(Section~\ref{ab:compatibility}), and the impact of key hyper-parameters~(Section~\ref{ab:parameters}).

\subsection{Sequence Parallel}
\label{ab:sp}
As discussed in Section~\ref{sec:parallel}, our dynamic programming (DP) algorithm requires $\mathcal{O}(LM)$ layer forward passes. 
To assess the time overhead, we conducted experiments on LLaMA3-70B using two NVIDIA A800 GPUs.
Without any parallel strategy, a single DP run to filter half of the layers takes approximately 2.5 seconds, whereas a single round of verification only takes about 0.1 seconds. 
After implementing our parallel strategy, the time for a single DP run is reduced to 0.14 seconds, comparable to the time for a single verification. 
This significantly reduces the additional latency introduced by layer optimization.

We further analyzed the per-query latency distribution of each stage, as illustrated in Figure~\ref{fig:proof}c. The results show that the latency proportion of layer optimization is significantly reduced with the parallel strategy. Additionally, with a lower update frequency, the extra update latency of \ours becomes almost negligible.

\subsection{Model Size Scaling Laws}
\label{ab:compatibility}
To assess the scalability of \ours, we evaluated its performance across a range of model sizes, including LLaMA2-13B and LLaMA3.1-405B, in addition to LLaMA3-8B and LLaMA3-70B. 
For LLaMA2-13B, the model was deployed on an A800 GPU using FP16 precision, while for LLaMA3.1-405B, we used INT8 quantization~\citep{dettmers-2022-int8} to deploy it on a single node with 8 A800 GPUs.

As illustrated in Figure~\ref{fig:different-model}, the speedup increases with model size across various tasks. 
Specifically, on the MT-bench, speedups range from $1.24\times$ for LLaMA3-8B to $1.73\times$ for LLaMA3.1-405B. 
For the GSM8K benchmark, speedups increase from $1.26\times$ to $1.81\times$, while on the Natural Questions benchmark, speedups range from $1.27\times$ to $1.82\times$. 
These results indicate that larger models exhibit enhanced layer sparsity, enabling \ours to leverage its capabilities more effectively and achieve greater speedups.

\subsection{Key Hyper-Parameters}
\label{ab:parameters}
We show the effect of key hyper-parameters on the acceleration benefits of \ours, where all experiments were performed using the LLaMA3-70B model on MT-Bench.

\subsubsection{Number of Skipped Layers} 
\label{sec:lsr}
Layer sparsity allows intermediate layers to be skipped, but the number of skipped layers directly influences performance. 
Adjusting this parameter involves a trade-off between draft quality and efficiency, both of which significantly impact the speedup. 

As shown in Figure~\ref{fig:layer-skip-ratio}, for LLaMA3-70B, which consists of 80 layers, the speedup increases as the number of skipped layers rises, reaching an optimal value of $1.64\times$ when 44 layers are skipped. Beyond this point, the benefits of a longer average acceptance length are outweighed by the increased cost of generating high-quality drafts, resulting in a decline in speedup.

\subsubsection{Layer Optimization Interval}
Performing layer optimization after every verification step is computationally expensive, as noted in Section~\ref{sec:lof}. 
Extending the Layer Optimization Interval (LOI) reduces the additional delays introduced by dynamic programming (DP) while having only a minor impact on the average acceptance length $\tau$. 

As illustrated in Figure~\ref{fig:layer-opt-interval}, the speedup initially increases with the LOI but begins to decline as the interval grows beyond 128. 
This decline is caused by a significant drop in $\tau$, which negatively impacts overall speedup.

\subsubsection{Draft-Exiting Threshold}
To balance draft efficiency and cost, skipping 40\% to 60\% of layers achieves an optimal trade-off, as noted in Section~\ref{sec:lsr}. 
However, the cost of a single draft remains high, necessitating a sufficiently high acceptance rate to maximize speedup. 

EAGLE-2~\citep{li-2024-eagle2} suggests leveraging the draft model's confidence score to approximate the acceptance rate. By tuning the Draft-Exiting Threshold (DET), we can control the acceptance rate to optimize acceleration. 

As shown in Figure~\ref{fig:draft-existing-threshold}, adjusting the DET around 0.7 results in the highest speedup. Even with higher DET values, high speedup is maintained, demonstrating the robustness of this parameter for achieving acceleration gains.

\section{Conclusion}
In this paper, we propose \textbf{CLaSp}, a novel self-speculative decoding framework that adaptively adjusts the layer-skipping strategy based on context. 
We discover the potential of context-aware layer sparsity for generating high-quality drafts. 
Leveraging this insight, \ours performs layer optimization before each draft stage with minimal additional latency, significantly increasing the decoding efficiency. 
Through extensive experiments across diverse text generation tasks, we demonstrated that \ours achieves consistent speedups of $\bm{1.3\times\sim1.7\times}$ over vanilla autoregressive decoding.
Furthermore, detailed analysis reveals that \ours generalizes well to different models and tasks. 
For future work, we aim to explore ways to better leverage the layer sparsity of LLMs to further reduce inference latency in larger models.

\section*{Limitations}
The \ours framework dynamically adjusts the layer-skipping strategy based on context, making the self-speculative decoding process of LLMs more efficient. 
However, certain limitations exist. 
Our experiments are conducted solely on NVIDIA A800 GPUs with 80GB of memory and limited to LLaMA series models, leaving the potential of more powerful hardware and other models unexplored. 
Additionally, while \ours can function alongside many existing speculative decoding innovations, we do not investigate these integrations. 
We believe that addressing these limitations and exploring such combinations in future research could lead to significant advancements.

\section*{Acknowledgments}
Min Yang was supported by National Key Research and Development Program of China (2022YFF0902100), National Natural Science Foundation of China (Grant No. 62376262), the Natural Science Foundation of Guangdong Province of China (2024A1515030166, 2025B1515020032), Shenzhen Science and Technology Innovation Program (KQTD20190929172835662).

\bibliography{custom}

\clearpage
\appendix





\end{document}

%% file: algs/clasp_dp.tex
\begin{algorithm}[t]
    \footnotesize
    \caption{\small \ours Skip Layer Strategy}
    \DontPrintSemicolon
    \label{alg:CLaSp}
    \KwIn{Num hidden layers $L$, num skip layers $M$, hidden states $X = \{x_0, x_1, ..., x_{L-1}\}$, DecoderLayer $f_i$, hidden size $d$}
    \KwOut{The optimal skipped layer set $\mathcal{S}$}
    $g \gets \text{zeros}(L + 1, M + 1, d)$, $g[0, 0] \gets x_0$ \\
    \ccmt{\textsc{//}~{\textit{Dynamic programming}}} \\
    \For {$i=1$ \KwTo $L+1$}{
        $g[i, 0] \gets x_i$ \\
        $\ell \gets \min(i - 1, M)$ \\
        $\mathcal{G} \gets f_{i - 1}(g[i - 1, 1: \ell + 1])$ \\
        $\mathcal{F} \gets \text{norm}(\text{cat}(\mathcal{G}, g[i - 1, : \ell]))$ \\
            $\sigma \gets \mathcal{F} \cdot \text{norm}(x_i)$ \\
            \If {$\sigma[:\ell] > \sigma[\ell:]$} {
                $g[i][1:\ell + 1] \gets \mathcal{G}$ \\
            } 
            \Else {
                $g[i][1:\ell + 1] \gets g[i - 1, : \ell]$ \\
            }
        \If {$i \le M$} {
            $g[i, i] \gets g[i - 1, i - 1]$ \\
        } 
    }
    $\mathcal{S} \gets \text{zeros}(L)$ \\
    \ccmt{\textsc{//}~{\textit{Backtracking optimal skipped layer set $\mathcal{S}$}}} \\
    \While {$i > 0 \; \textbf{and} \; j > 0$} {
        \If {$g[i, j] = g[i - 1, j - 1]$} {
            $S[i - 1] \gets 1$ \\
            $j \gets j - 1$ \\
        }
        $i \gets i - 1$ \\
    }
    \Return{$\mathcal{S}$}
\end{algorithm}

%% file: tabs/main.tex
\begin{table*}[!t]
\centering
\small
\vspace{-0.6cm}
\setlength{\tabcolsep}{1.4mm}
\resizebox{\linewidth}{!}{
\begin{tabular}{llcccccccccccc|cc}
\toprule
\multirow{2}{*}{\textbf{Models}} & \multirow{2}{*}{\textbf{Methods}}
& \multicolumn{2}{c}{\textbf{MT-bench}} & \multicolumn{2}{c}{\textbf{WMT14}} & \multicolumn{2}{c}{\textbf{CNN/DM}} & \multicolumn{2}{c}{\textbf{NQ}} & \multicolumn{2}{c}{\textbf{GSM8K}} & \multicolumn{2}{c}{\textbf{DPR}} & \multirow{2}{*}{\begin{tabular}[c]{@{}c@{}}\textbf{Overall} \\\textbf{Speedup}\end{tabular}} \\ 
\cmidrule(lr){3-4} \cmidrule(lr){5-6} \cmidrule(lr){7-8} \cmidrule(lr){9-10} \cmidrule(lr){11-12} \cmidrule(lr){13-14}
& &$\tau$ & Speedup &$\tau$ & Speedup &$\tau$ & Speedup &$\tau$ & Speedup &$\tau$ & Speedup &$\tau$ & Speedup \\ 
\midrule
\multicolumn{15}{c}{\bf{Greedy Setting: Temperature=0}} \\
\midrule
\multirow{4}{*}{LLaMA-3-70B} & 
\textsc{Autoregressive} & 1.00 & 1.00$\times$ & 1.00 & 1.00$\times$ & 1.00 & 1.00$\times$ & 1.00 & 1.00$\times$ & 1.00 & 1.00$\times$ & 1.00 & 1.00$\times$ & 1.00$\times$ \\
& \textsc{Self-SD} & 2.57 & 1.38$\times$ & 4.10 & 1.55$\times$ & 5.46 & 1.57$\times$ & 2.60 & 1.42$\times$ & 3.10 & 1.49$\times$ & 3.59 & 1.43$\times$ & 1.47$\times$ \\
& \textsc{SWIFT} & 3.13 & 1.26$\times$ & 2.90 & 1.27$\times$ & 3.93 & 1.35$\times$ & 3.21 & 1.29$\times$ & 2.86 & 1.27$\times$ & 3.31 & 1.26$\times$ & 1.28$\times$ \\
& \orange{\textsc{\ours}} & \orange{4.55} & \orange{\textbf{1.64$\times$}} & \orange{5.81} & \orange{\textbf{1.69$\times$}} & \orange{7.19} & \orange{\textbf{1.66$\times$}} & \orange{5.37} & \orange{\textbf{1.72$\times$}} & \orange{6.77} & \orange{\textbf{1.75$\times$}} & \orange{4.05} & \orange{\textbf{1.56$\times$}} & \orange{\textbf{1.67$\times$}} \\
\midrule
\multirow{4}{*}{\begin{tabular}[c]{@{}c@{}}LLaMA-3-70B\\-Chat\end{tabular}} &
\textsc{Autoregressive} & 1.00 & 1.00$\times$ & 1.00 & 1.00$\times$ & 1.00 & 1.00$\times$ & 1.00 & 1.00$\times$ & 1.00 & 1.00$\times$ & 1.00 & 1.00$\times$ & 1.00$\times$ \\
& \textsc{Self-SD} & 1.40 & 1.23$\times$ & 2.27 & 1.33$\times$ & 1.50 & 1.24$\times$ & 1.59 & 1.26$\times$ & 3.00 & 1.40$\times$ & 2.56 & 1.37$\times$ & 1.31$\times$ \\
& \textsc{SWIFT} & 4.41 & 1.15$\times$ & 5.54 & 1.27$\times$ & 4.52 & 1.22$\times$ & 4.83 & 1.20$\times$ & 6.19 & 1.31$\times$ & 5.97 & 1.33$\times$ & 1.25$\times$ \\
& \orange{\textsc{\ours}} & \orange{2.61} & \orange{\textbf{1.35$\times$}} & \orange{4.72} & \orange{\textbf{1.51$\times$}} & \orange{3.48} & \orange{\textbf{1.39$\times$}} & \orange{3.32} & \orange{\textbf{1.39$\times$}} & \orange{5.28} & \orange{\textbf{1.53$\times$}} & \orange{5.61} & \orange{\textbf{1.54$\times$}} & \orange{\textbf{1.45$\times$}} \\
\midrule
\multirow{4}{*}{LLaMA-3-8B} &
\textsc{Autoregressive} & 1.00 & 1.00$\times$ & 1.00 & 1.00$\times$ & 1.00 & 1.00$\times$ & 1.00 & 1.00$\times$ & 1.00 & 1.00$\times$ & 1.00 & 1.00$\times$ & 1.00$\times$ \\
& \textsc{Self-SD} & 1.28 & 1.07$\times$ & 1.35 & 1.13$\times$ & 1.73 & 1.17$\times$ & 1.45 & 1.13$\times$ & 1.44 & 1.15$\times$ & 2.33 & 1.21$\times$ & 1.14$\times$ \\
& \textsc{SWIFT} & 2.75 & 1.07$\times$ & 2.51 & 1.09$\times$ & 2.76 & 1.13$\times$ & 2.91 & 1.13$\times$ & 2.72 & 1.10$\times$ & 2.96 & 1.11$\times$ & 1.11$\times$ \\
& \orange{\textsc{\ours}} & \orange{3.68} & \orange{\textbf{1.24$\times$}} & \orange{4.14} & \orange{\textbf{1.23$\times$}} & \orange{6.22} & \orange{\textbf{1.22$\times$}} & \orange{4.03} & \orange{\textbf{1.27$\times$}} & \orange{5.26} & \orange{\textbf{1.26$\times$}} & \orange{4.17} & \orange{\textbf{1.22$\times$}} & \orange{\textbf{1.24$\times$}} \\
\midrule
\multicolumn{15}{c}{\bf{Non-Greedy Setting: Temperature=1}} \\
\midrule
\multirow{4}{*}{LLaMA-3-70B} & 
\textsc{Autoregressive} & 1.00 & 1.00$\times$ & 1.00 & 1.00$\times$ & 1.00 & 1.00$\times$ & 1.00 & 1.00$\times$ & 1.00 & 1.00$\times$ & 1.00 & 1.00$\times$ & 1.00$\times$ \\
& \textsc{Self-SD} & 1.64 & 1.23$\times$ & 2.53 & 1.39$\times$ & 3.61 & 1.43$\times$ & 1.53 & 1.24$\times$ & 2.01 & 1.33$\times$ & 2.17 & 1.24$\times$ & 1.31$\times$ \\
& \textsc{SWIFT} & 2.06 & 1.10$\times$ & 1.96 & 1.08$\times$ & 1.97 & 1.09$\times$ & 1.97 & 1.08$\times$ & 1.98 & 1.09$\times$ & 2.01 & 1.07$\times$ & 1.09$\times$ \\
& \orange{\textsc{\ours}} & \orange{3.13} & \orange{\textbf{1.49$\times$}} & \orange{3.33} & \orange{\textbf{1.50$\times$}} & \orange{5.38} & \orange{\textbf{1.54$\times$}} & \orange{3.56} & \orange{\textbf{1.54$\times$}} & \orange{4.32} & \orange{\textbf{1.59$\times$}} & \orange{2.51} & \orange{\textbf{1.36$\times$}} & \orange{\textbf{1.50$\times$}} \\
\midrule
\multirow{4}{*}{\begin{tabular}[c]{@{}c@{}}LLaMA-3-70B\\-Chat\end{tabular}} &
\textsc{Autoregressive} & 1.00 & 1.00$\times$ & 1.00 & 1.00$\times$ & 1.00 & 1.00$\times$ & 1.00 & 1.00$\times$ & 1.00 & 1.00$\times$ & 1.00 & 1.00$\times$ & 1.00$\times$ \\
& \textsc{Self-SD} & 1.15 & 1.14$\times$ & 2.01 & 1.23$\times$ & 1.19 & 1.15$\times$ & 1.21 & 1.17$\times$ & 1.97 & 1.34$\times$ & 1.71 & 1.26$\times$ & 1.22$\times$ \\
& \textsc{SWIFT} & 2.68 & 0.96$\times$ & 2.64 & 0.99$\times$ & 2.67 & 0.98$\times$ & 2.62 & 0.99$\times$ & 2.79 & 1.01$\times$ & 2.76 & 1.04$\times$ & 1.00$\times$ \\
& \orange{\textsc{\ours}} & \orange{1.96} & \orange{\textbf{1.28$\times$}} & \orange{3.90} & \orange{\textbf{1.45$\times$}} & \orange{2.32} & \orange{\textbf{1.29$\times$}} & \orange{2.28} & \orange{\textbf{1.30$\times$}} & \orange{4.40} & \orange{\textbf{1.47$\times$}} & \orange{4.03} & \orange{\textbf{1.43$\times$}} & \orange{\textbf{1.37$\times$}} \\
\midrule
\multirow{4}{*}{LLaMA-3-8B} &
\textsc{Autoregressive} & 1.00 & 1.00$\times$ & 1.00 & 1.00$\times$ & 1.00 & 1.00$\times$ & 1.00 & 1.00$\times$ & 1.00 & 1.00$\times$ & 1.00 & 1.00$\times$ & 1.00$\times$ \\
& \textsc{Self-SD} & 0.98 & 0.89$\times$ & 1.01 & 0.94$\times$ & 1.36 & 1.02$\times$ & 1.09 & 0.92$\times$ & 1.09 & 0.96$\times$ & 1.82 & 1.03$\times$ & 0.96$\times$ \\
& \textsc{SWIFT} & 1.90 & 0.80$\times$ & 1.92 & 0.85$\times$ & 1.85 & 0.83$\times$ & 1.97 & 0.84$\times$ & 1.95 & 0.83$\times$ & 1.90 & 0.80$\times$ & 0.83$\times$ \\
& \orange{\textsc{\ours}} & \orange{2.62} & \orange{\textbf{1.11$\times$}} & \orange{2.78} & \orange{\textbf{1.08$\times$}} & \orange{4.26} & \orange{\textbf{1.11$\times$}} & \orange{2.70} & \orange{\textbf{1.08$\times$}} & \orange{3.76} & \orange{\textbf{1.10$\times$}} & \orange{2.35} & \orange{\textbf{1.02$\times$}} & \orange{\textbf{1.08$\times$}} \\
\bottomrule
\end{tabular}}
\caption{Comparison between \ours and prior plug-and-play methods. We report the average acceptance length $\tau$ and speedup ratio under greedy (Temperature=0) and non-greedy (Temperature=1) settings. \textbf{Bold} numbers denotes the best Speedup.}
\label{tab:main-exp}
\end{table*}

%% file: acl_latex.bbl
\begin{thebibliography}{46}
\providecommand{\natexlab}[1]{#1}

\bibitem[{pan(2024)}]{pan-2024-lisa}
 2024.
\newblock \href {https://arxiv.org/abs/2403.17919} {Lisa: Layerwise importance sampling for memory-efficient large language model fine-tuning}.
\newblock \emph{arXiv preprint arXiv:2403.17919}.

\bibitem[{Achiam et~al.(2023)Achiam, Adler, Agarwal, Ahmad, Akkaya, Aleman, Almeida, Altenschmidt, Altman, Anadkat et~al.}]{achiam-2023-gpt4}
Josh Achiam, Steven Adler, Sandhini Agarwal, Lama Ahmad, Ilge Akkaya, Florencia~Leoni Aleman, Diogo Almeida, Janko Altenschmidt, Sam Altman, Shyamal Anadkat, et~al. 2023.
\newblock \href {https://arxiv.org/abs/2303.08774} {Gpt-4 technical report}.
\newblock \emph{arXiv preprint arXiv:2303.08774}.

\bibitem[{Agrawal et~al.(2023)Agrawal, Panwar, Mohan, Kwatra, Gulavani, and Ramjee}]{agrawal-2023-sarathi}
Amey Agrawal, Ashish Panwar, Jayashree Mohan, Nipun Kwatra, Bhargav~S Gulavani, and Ramachandran Ramjee. 2023.
\newblock \href {https://arxiv.org/abs/2308.16369} {Sarathi: Efficient llm inference by piggybacking decodes with chunked prefills}.
\newblock \emph{arXiv preprint arXiv:2308.16369}.

\bibitem[{Anil et~al.(2023)Anil, Dai, Firat, Johnson, Lepikhin, Passos, Shakeri, Taropa, Bailey, Chen et~al.}]{anil-2023-palm}
Rohan Anil, Andrew~M Dai, Orhan Firat, Melvin Johnson, Dmitry Lepikhin, Alexandre Passos, Siamak Shakeri, Emanuel Taropa, Paige Bailey, Zhifeng Chen, et~al. 2023.
\newblock \href {https://arxiv.org/abs/2305.10403} {Palm 2 technical report}.
\newblock \emph{arXiv preprint arXiv:2305.10403}.

\bibitem[{Brown et~al.(2020)Brown, Mann, Ryder, Subbiah, Kaplan, Dhariwal, Neelakantan, Shyam, Sastry, Askell, Agarwal, Herbert-Voss, Krueger, Henighan, Child, Ramesh, Ziegler, Wu, Winter, Hesse, Chen, Sigler, Litwin, Gray, Chess, Clark, Berner, McCandlish, Radford, Sutskever, and Amodei}]{brown-2020-language}
Tom Brown, Benjamin Mann, Nick Ryder, Melanie Subbiah, Jared~D Kaplan, Prafulla Dhariwal, Arvind Neelakantan, Pranav Shyam, Girish Sastry, Amanda Askell, Sandhini Agarwal, Ariel Herbert-Voss, Gretchen Krueger, Tom Henighan, Rewon Child, Aditya Ramesh, Daniel Ziegler, Jeffrey Wu, Clemens Winter, Chris Hesse, Mark Chen, Eric Sigler, Mateusz Litwin, Scott Gray, Benjamin Chess, Jack Clark, Christopher Berner, Sam McCandlish, Alec Radford, Ilya Sutskever, and Dario Amodei. 2020.
\newblock \href {https://proceedings.neurips.cc/paper_files/paper/2020/file/1457c0d6bfcb4967418bfb8ac142f64a-Paper.pdf} {Language models are few-shot learners}.
\newblock In \emph{Advances in Neural Information Processing Systems}, volume~33, pages 1877--1901. Curran Associates, Inc.

\bibitem[{Burton(1985)}]{burton-1985-speculative}
F.~Warren Burton. 1985.
\newblock \href {https://doi.org/10.1109/TC.1985.6312218} {Speculative computation, parallelism, and functional programming}.
\newblock \emph{IEEE Transactions on Computers}, C-34(12):1190--1193.

\bibitem[{Cai et~al.(2024)Cai, Li, Geng, Peng, Lee, Chen, and Dao}]{cai-2024-medusa}
Tianle Cai, Yuhong Li, Zhengyang Geng, Hongwu Peng, Jason~D. Lee, Deming Chen, and Tri Dao. 2024.
\newblock \href {https://openreview.net/forum?id=PEpbUobfJv} {Medusa: Simple {LLM} inference acceleration framework with multiple decoding heads}.
\newblock In \emph{Forty-first International Conference on Machine Learning, {ICML} 2024, Vienna, Austria, July 21-27, 2024}. OpenReview.net.

\bibitem[{Chen et~al.(2023)Chen, Borgeaud, Irving, Lespiau, Sifre, and Jumper}]{chen-2023-accelerating}
Charlie Chen, Sebastian Borgeaud, Geoffrey Irving, Jean-Baptiste Lespiau, Laurent Sifre, and John Jumper. 2023.
\newblock \href {https://arxiv.org/abs/2302.01318} {Accelerating large language model decoding with speculative sampling}.
\newblock \emph{arXiv preprint arXiv:2302.01318}.

\bibitem[{Chen et~al.(2024)Chen, May, Svirschevski, Huang, Ryabinin, Jia, and Chen}]{chen-2024-sequoia}
Zhuoming Chen, Avner May, Ruslan Svirschevski, Yuhsun Huang, Max Ryabinin, Zhihao Jia, and Beidi Chen. 2024.
\newblock \href {https://arxiv.org/abs/2402.12374} {Sequoia: Scalable, robust, and hardware-aware speculative decoding}.
\newblock \emph{arXiv preprint arXiv:2402.12374}.

\bibitem[{Cobbe et~al.(2021)Cobbe, Kosaraju, Bavarian, Chen, Jun, Kaiser, Plappert, Tworek, Hilton, Nakano et~al.}]{cobbe-2021-training}
Karl Cobbe, Vineet Kosaraju, Mohammad Bavarian, Mark Chen, Heewoo Jun, Lukasz Kaiser, Matthias Plappert, Jerry Tworek, Jacob Hilton, Reiichiro Nakano, et~al. 2021.
\newblock \href {https://arxiv.org/abs/2110.14168} {Training verifiers to solve math word problems}.
\newblock \emph{arXiv preprint arXiv:2110.14168}.

\bibitem[{Corro et~al.(2023)Corro, Giorno, Agarwal, Yu, Awadallah, and Mukherjee}]{luciano-2023-skipdecode}
Luciano~Del Corro, Allie~Del Giorno, Sahaj Agarwal, Bin Yu, Ahmed Awadallah, and Subhabrata Mukherjee. 2023.
\newblock \href {https://arxiv.org/abs/2307.02628} {Skipdecode: Autoregressive skip decoding with batching and caching for efficient llm inference}.
\newblock \emph{arXiv preprint arXiv:2307.02628}.

\bibitem[{Dettmers et~al.(2022)Dettmers, Lewis, Belkada, and Zettlemoyer}]{dettmers-2022-int8}
Tim Dettmers, Mike Lewis, Younes Belkada, and Luke Zettlemoyer. 2022.
\newblock \href {https://proceedings.neurips.cc/paper_files/paper/2022/file/c3ba4962c05c49636d4c6206a97e9c8a-Paper-Conference.pdf} {Gpt3.int8(): 8-bit matrix multiplication for transformers at scale}.
\newblock In \emph{Advances in Neural Information Processing Systems}, volume~35, pages 30318--30332. Curran Associates, Inc.

\bibitem[{Du et~al.(2024)Du, Jiang, Xu, Wu, Yu, Li, Li, Xu, Nie, Tu, and You}]{du-2024-glide}
Cunxiao Du, Jing Jiang, Yuanchen Xu, Jiawei Wu, Sicheng Yu, Yongqi Li, Shenggui Li, Kai Xu, Liqiang Nie, Zhaopeng Tu, and Yang You. 2024.
\newblock \href {https://openreview.net/forum?id=mk8oRhox2l} {Glide with a cape: {A} low-hassle method to accelerate speculative decoding}.
\newblock In \emph{Forty-first International Conference on Machine Learning, {ICML} 2024, Vienna, Austria, July 21-27, 2024}. OpenReview.net.

\bibitem[{Dubey et~al.(2024)Dubey, Jauhri, Pandey, Kadian, Al-Dahle, Letman, Mathur, Schelten, Yang, Fan et~al.}]{dubey-2024-llama3}
Abhimanyu Dubey, Abhinav Jauhri, Abhinav Pandey, Abhishek Kadian, Ahmad Al-Dahle, Aiesha Letman, Akhil Mathur, Alan Schelten, Amy Yang, Angela Fan, et~al. 2024.
\newblock \href {https://arxiv.org/abs/2407.21783} {The llama 3 herd of models}.
\newblock \emph{arXiv preprint arXiv:2407.21783}.

\bibitem[{Elhoushi et~al.(2024)Elhoushi, Shrivastava, Liskovich, Hosmer, Wasti, Lai, Mahmoud, Acun, Agarwal, Roman, Aly, Chen, and Wu}]{elhoushi-2024-layerskip}
Mostafa Elhoushi, Akshat Shrivastava, Diana Liskovich, Basil Hosmer, Bram Wasti, Liangzhen Lai, Anas Mahmoud, Bilge Acun, Saurabh Agarwal, Ahmed Roman, Ahmed Aly, Beidi Chen, and Carole-Jean Wu. 2024.
\newblock \href {https://doi.org/10.18653/v1/2024.acl-long.681} {{L}ayer{S}kip: Enabling early exit inference and self-speculative decoding}.
\newblock In \emph{Proceedings of the 62nd Annual Meeting of the Association for Computational Linguistics (Volume 1: Long Papers)}, pages 12622--12642, Bangkok, Thailand. Association for Computational Linguistics.

\bibitem[{Fan et~al.(2020)Fan, Grave, and Joulin}]{angela-2020-reducing}
Angela Fan, Edouard Grave, and Armand Joulin. 2020.
\newblock \href {https://openreview.net/forum?id=SylO2yStDr} {Reducing transformer depth on demand with structured dropout}.
\newblock In \emph{8th International Conference on Learning Representations, {ICLR} 2020, Addis Ababa, Ethiopia, April 26-30, 2020}. OpenReview.net.

\bibitem[{Glavas et~al.(2024)Glavas, Chataoui, Regol, Jabbour, Valkanas, Oreshkin, and Coates}]{glavas-2024-dynamic}
Theodore Glavas, Joud Chataoui, Florence Regol, Wassim Jabbour, Antonios Valkanas, Boris~N. Oreshkin, and Mark Coates. 2024.
\newblock \href {https://arxiv.org/abs/2410.20022} {Dynamic layer selection in decoder-only transformers}.
\newblock \emph{arXiv preprint arXiv:2410.20022}.

\bibitem[{Gugger et~al.(2022)Gugger, Debut, Wolf, Schmid, Mueller, Mangrulkar, Sun, and Bossan}]{gugger-2022-accelerate}
Sylvain Gugger, Lysandre Debut, Thomas Wolf, Philipp Schmid, Zachary Mueller, Sourab Mangrulkar, Marc Sun, and Benjamin Bossan. 2022.
\newblock Accelerate: Training and inference at scale made simple, efficient and adaptable.
\newblock \url{https://github.com/huggingface/accelerate}.

\bibitem[{He et~al.(2024)He, Zhong, Cai, Lee, and He}]{he-2024-rest}
Zhenyu He, Zexuan Zhong, Tianle Cai, Jason Lee, and Di~He. 2024.
\newblock \href {https://doi.org/10.18653/v1/2024.naacl-long.88} {{REST}: Retrieval-based speculative decoding}.
\newblock In \emph{Proceedings of the 2024 Conference of the North American Chapter of the Association for Computational Linguistics: Human Language Technologies (Volume 1: Long Papers)}, pages 1582--1595, Mexico City, Mexico. Association for Computational Linguistics.

\bibitem[{Hennessy and Patterson(2012)}]{hennessy-2017-computer}
John~L. Hennessy and David~A. Patterson. 2012.
\newblock \emph{Computer Architecture: A Quantitative Approach}, 5 edition.
\newblock Morgan Kaufmann, Amsterdam.

\bibitem[{Kaplan et~al.(2020)Kaplan, McCandlish, Henighan, Brown, Chess, Child, Gray, Radford, Wu, and Amodei}]{kaplan-2020-scaling}
Jared Kaplan, Sam McCandlish, Tom Henighan, Tom~B Brown, Benjamin Chess, Rewon Child, Scott Gray, Alec Radford, Jeffrey Wu, and Dario Amodei. 2020.
\newblock \href {https://arxiv.org/abs/2001.08361} {Scaling laws for neural language models}.
\newblock \emph{arXiv preprint arXiv:2001.08361}.

\bibitem[{Karpukhin et~al.(2020)Karpukhin, Oguz, Min, Lewis, Wu, Edunov, Chen, and Yih}]{karpukhin-2020-dense}
Vladimir Karpukhin, Barlas Oguz, Sewon Min, Patrick Lewis, Ledell Wu, Sergey Edunov, Danqi Chen, and Wen-tau Yih. 2020.
\newblock \href {https://doi.org/10.18653/v1/2020.emnlp-main.550} {Dense passage retrieval for open-domain question answering}.
\newblock In \emph{Proceedings of the 2020 Conference on Empirical Methods in Natural Language Processing (EMNLP)}, pages 6769--6781, Online. Association for Computational Linguistics.

\bibitem[{Kwiatkowski et~al.(2019)Kwiatkowski, Palomaki, Redfield, Collins, Parikh, Alberti, Epstein, Polosukhin, Devlin, Lee, Toutanova, Jones, Kelcey, Chang, Dai, Uszkoreit, Le, and Petrov}]{kwiatkowski-2019-natural}
Tom Kwiatkowski, Jennimaria Palomaki, Olivia Redfield, Michael Collins, Ankur Parikh, Chris Alberti, Danielle Epstein, Illia Polosukhin, Jacob Devlin, Kenton Lee, Kristina Toutanova, Llion Jones, Matthew Kelcey, Ming-Wei Chang, Andrew~M. Dai, Jakob Uszkoreit, Quoc Le, and Slav Petrov. 2019.
\newblock \href {https://doi.org/10.1162/tacl_a_00276} {Natural questions: A benchmark for question answering research}.
\newblock \emph{Transactions of the Association for Computational Linguistics}, 7:452--466.

\bibitem[{Leviathan et~al.(2023)Leviathan, Kalman, and Matias}]{leviathan-2023-fast}
Yaniv Leviathan, Matan Kalman, and Yossi Matias. 2023.
\newblock \href {https://proceedings.mlr.press/v202/leviathan23a.html} {Fast inference from transformers via speculative decoding}.
\newblock In \emph{Proceedings of the 40th International Conference on Machine Learning}, volume 202 of \emph{Proceedings of Machine Learning Research}, pages 19274--19286. PMLR.

\bibitem[{Li et~al.(2024{\natexlab{a}})Li, Wei, Zhang, and Zhang}]{li-2024-eagle2}
Yuhui Li, Fangyun Wei, Chao Zhang, and Hongyang Zhang. 2024{\natexlab{a}}.
\newblock \href {https://arxiv.org/abs/2406.16858} {{EAGLE-2}: Faster inference of language models with dynamic draft trees}.
\newblock \emph{arXiv preprint arXiv:2406.16858}.

\bibitem[{Li et~al.(2024{\natexlab{b}})Li, Wei, Zhang, and Zhang}]{li-2024-eagle}
Yuhui Li, Fangyun Wei, Chao Zhang, and Hongyang Zhang. 2024{\natexlab{b}}.
\newblock \href {https://openreview.net/forum?id=1NdN7eXyb4} {{EAGLE:} speculative sampling requires rethinking feature uncertainty}.
\newblock In \emph{Forty-first International Conference on Machine Learning, {ICML} 2024, Vienna, Austria, July 21-27, 2024}. OpenReview.net.

\bibitem[{Liu et~al.(2024)Liu, Tang, Liu, Ni, Han, and Wang}]{liu-2024-kangaroo}
Fangcheng Liu, Yehui Tang, Zhenhua Liu, Yunsheng Ni, Kai Han, and Yunhe Wang. 2024.
\newblock \href {https://arxiv.org/abs/2404.18911} {Kangaroo: Lossless self-speculative decoding via double early exiting}.
\newblock \emph{arXiv preprint arXiv:2404.18911}.

\bibitem[{Liu et~al.(2023)Liu, Wang, Dao, Zhou, Yuan, Song, Shrivastava, Zhang, Tian, Re, and Chen}]{liu-2023-dejavu}
Zichang Liu, Jue Wang, Tri Dao, Tianyi Zhou, Binhang Yuan, Zhao Song, Anshumali Shrivastava, Ce~Zhang, Yuandong Tian, Christopher Re, and Beidi Chen. 2023.
\newblock \href {https://proceedings.mlr.press/v202/liu23am.html} {Deja vu: Contextual sparsity for efficient {LLM}s at inference time}.
\newblock In \emph{Proceedings of the 40th International Conference on Machine Learning}, volume 202 of \emph{Proceedings of Machine Learning Research}, pages 22137--22176. PMLR.

\bibitem[{Miao et~al.(2024)Miao, Oliaro, Zhang, Cheng, Wang, Zhang, Wong, Zhu, Yang, Shi, Shi, Chen, Arfeen, Abhyankar, and Jia}]{miao-2023-specinfer}
Xupeng Miao, Gabriele Oliaro, Zhihao Zhang, Xinhao Cheng, Zeyu Wang, Zhengxin Zhang, Rae Ying~Yee Wong, Alan Zhu, Lijie Yang, Xiaoxiang Shi, Chunan Shi, Zhuoming Chen, Daiyaan Arfeen, Reyna Abhyankar, and Zhihao Jia. 2024.
\newblock \href {https://doi.org/10.1145/3620666.3651335} {Specinfer: Accelerating large language model serving with tree-based speculative inference and verification}.
\newblock In \emph{Proceedings of the 29th ACM International Conference on Architectural Support for Programming Languages and Operating Systems, Volume 3}, ASPLOS '24, page 932–949, New York, NY, USA. Association for Computing Machinery.

\bibitem[{Nallapati et~al.(2016)Nallapati, Zhou, dos Santos, Gulcehre, and Xiang}]{nallapati-2016-abstractive}
Ramesh Nallapati, Bowen Zhou, Cicero dos Santos, Caglar Gulcehre, and Bing Xiang. 2016.
\newblock \href {https://doi.org/10.18653/v1/K16-1028} {Abstractive text summarization using sequence-to-sequence {RNN}s and beyond}.
\newblock In \emph{Proceedings of the 20th {SIGNLL} Conference on Computational Natural Language Learning}, pages 280--290, Berlin, Germany. Association for Computational Linguistics.

\bibitem[{Patterson(2004)}]{patterson-2004-latency}
David~A. Patterson. 2004.
\newblock \href {https://doi.org/10.1145/1022594.1022596} {Latency lags bandwith}.
\newblock \emph{Commun. ACM}, 47(10):71–75.

\bibitem[{Reid et~al.(2024)Reid, Savinov, Teplyashin, Lepikhin, Lillicrap, Alayrac, Soricut, Lazaridou, Firat, Schrittwieser et~al.}]{reid-2024-gemini}
Machel Reid, Nikolay Savinov, Denis Teplyashin, Dmitry Lepikhin, Timothy Lillicrap, Jean-baptiste Alayrac, Radu Soricut, Angeliki Lazaridou, Orhan Firat, Julian Schrittwieser, et~al. 2024.
\newblock \href {https://arxiv.org/abs/2403.05530} {Gemini 1.5: Unlocking multimodal understanding across millions of tokens of context}.
\newblock \emph{arXiv preprint arXiv:2403.05530}.

\bibitem[{Saxena(2023)}]{saxena-2023-prompt}
Apoorv Saxena. 2023.
\newblock \href {https://github.com/apoorvumang/prompt-lookup-decoding/} {Prompt lookup decoding}.

\bibitem[{Shazeer(2019)}]{shazeer-2019-fast}
Noam Shazeer. 2019.
\newblock \href {https://arxiv.org/abs/1911.02150} {Fast transformer decoding: One write-head is all you need}.
\newblock \emph{arXiv preprint arXiv:1911.02150}.

\bibitem[{Sun et~al.(2024)Sun, Chen, Yang, Tian, and Chen}]{sun-2024-triforce}
Hanshi Sun, Zhuoming Chen, Xinyu Yang, Yuandong Tian, and Beidi Chen. 2024.
\newblock \href {https://arxiv.org/abs/2404.11912} {Triforce: Lossless acceleration of long sequence generation with hierarchical speculative decoding}.
\newblock \emph{arXiv preprint arXiv:2404.11912}.

\bibitem[{Svirschevski et~al.(2024)Svirschevski, May, Chen, Chen, Jia, and Ryabinin}]{svirschevski-2024-specexec}
Ruslan Svirschevski, Avner May, Zhuoming Chen, Beidi Chen, Zhihao Jia, and Max Ryabinin. 2024.
\newblock \href {https://arxiv.org/abs/2406.02532} {Specexec: Massively parallel speculative decoding for interactive llm inference on consumer devices}.
\newblock \emph{arXiv preprint arXiv:2406.02532}.

\bibitem[{Touvron et~al.(2023{\natexlab{a}})Touvron, Lavril, Izacard, Martinet, Lachaux, Lacroix, Rozi{\`e}re, Goyal, Hambro, Azhar et~al.}]{touvron-2023-llama}
Hugo Touvron, Thibaut Lavril, Gautier Izacard, Xavier Martinet, Marie-Anne Lachaux, Timoth{\'e}e Lacroix, Baptiste Rozi{\`e}re, Naman Goyal, Eric Hambro, Faisal Azhar, et~al. 2023{\natexlab{a}}.
\newblock \href {https://arxiv.org/abs/2302.13971} {Llama: Open and efficient foundation language models}.
\newblock \emph{arXiv preprint arXiv:2302.13971}.

\bibitem[{Touvron et~al.(2023{\natexlab{b}})Touvron, Martin, Stone, Albert, Almahairi, Babaei, Bashlykov, Batra, Bhargava, Bhosale et~al.}]{touvron-2023-llama2}
Hugo Touvron, Louis Martin, Kevin Stone, Peter Albert, Amjad Almahairi, Yasmine Babaei, Nikolay Bashlykov, Soumya Batra, Prajjwal Bhargava, Shruti Bhosale, et~al. 2023{\natexlab{b}}.
\newblock \href {https://arxiv.org/abs/2307.09288} {Llama 2: Open foundation and fine-tuned chat models}.
\newblock \emph{arXiv preprint arXiv:2307.09288}.

\bibitem[{Xia et~al.(2023)Xia, Ge, Wang, Chen, Wei, and Sui}]{xia-2023-speculative}
Heming Xia, Tao Ge, Peiyi Wang, Si-Qing Chen, Furu Wei, and Zhifang Sui. 2023.
\newblock \href {https://doi.org/10.18653/v1/2023.findings-emnlp.257} {Speculative decoding: Exploiting speculative execution for accelerating seq2seq generation}.
\newblock In \emph{Findings of the Association for Computational Linguistics: EMNLP 2023}, pages 3909--3925, Singapore. Association for Computational Linguistics.

\bibitem[{Xia et~al.(2024{\natexlab{a}})Xia, Li, Zhang, Du, and Li}]{xia-2024-swift}
Heming Xia, Yongqi Li, Jun Zhang, Cunxiao Du, and Wenjie Li. 2024{\natexlab{a}}.
\newblock \href {https://arxiv.org/abs/2410.06916} {Swift: On-the-fly self-speculative decoding for llm inference acceleration}.
\newblock \emph{arXiv preprint arXiv:2410.06916}.

\bibitem[{Xia et~al.(2024{\natexlab{b}})Xia, Yang, Dong, Wang, Li, Ge, Liu, Li, and Sui}]{xia-2024-unlocking}
Heming Xia, Zhe Yang, Qingxiu Dong, Peiyi Wang, Yongqi Li, Tao Ge, Tianyu Liu, Wenjie Li, and Zhifang Sui. 2024{\natexlab{b}}.
\newblock \href {https://doi.org/10.18653/v1/2024.findings-acl.456} {Unlocking efficiency in large language model inference: A comprehensive survey of speculative decoding}.
\newblock In \emph{Findings of the Association for Computational Linguistics ACL 2024}, pages 7655--7671, Bangkok, Thailand and virtual meeting. Association for Computational Linguistics.

\bibitem[{Yang et~al.(2024)Yang, Yang, Hui, Zheng, Yu, Zhou, Li, Li, Liu, Huang et~al.}]{yang-2024-qwen2}
An~Yang, Baosong Yang, Binyuan Hui, Bo~Zheng, Bowen Yu, Chang Zhou, Chengpeng Li, Chengyuan Li, Dayiheng Liu, Fei Huang, et~al. 2024.
\newblock \href {https://arxiv.org/abs/2407.10671} {Qwen2 technical report}.
\newblock \emph{arXiv preprint arXiv:2407.10671}.

\bibitem[{Zhang et~al.(2024)Zhang, Wang, Li, Shou, Chen, Chen, and Mehrotra}]{zhang-2024-draft}
Jun Zhang, Jue Wang, Huan Li, Lidan Shou, Ke~Chen, Gang Chen, and Sharad Mehrotra. 2024.
\newblock \href {https://doi.org/10.18653/v1/2024.acl-long.607} {Draft{\&} verify: Lossless large language model acceleration via self-speculative decoding}.
\newblock In \emph{Proceedings of the 62nd Annual Meeting of the Association for Computational Linguistics (Volume 1: Long Papers)}, pages 11263--11282, Bangkok, Thailand. Association for Computational Linguistics.

\bibitem[{Zhang et~al.(2023)Zhang, Ding, Qi, Zhu, Long, and Zhou}]{zhang-2023-crash}
Kaiyan Zhang, Ning Ding, Biqing Qi, Xuekai Zhu, Xinwei Long, and Bowen Zhou. 2023.
\newblock \href {https://doi.org/10.18653/v1/2023.emnlp-main.597} {{CR}a{S}h: Clustering, removing, and sharing enhance fine-tuning without full large language model}.
\newblock In \emph{Proceedings of the 2023 Conference on Empirical Methods in Natural Language Processing}, pages 9612--9637, Singapore. Association for Computational Linguistics.

\bibitem[{Zhang and He(2020)}]{zhang-2020-accelerating}
Minjia Zhang and Yuxiong He. 2020.
\newblock \href {https://proceedings.neurips.cc/paper_files/paper/2020/file/a1140a3d0df1c81e24ae954d935e8926-Paper.pdf} {Accelerating training of transformer-based language models with progressive layer dropping}.
\newblock In \emph{Advances in Neural Information Processing Systems}, volume~33, pages 14011--14023. Curran Associates, Inc.

\bibitem[{Zheng et~al.(2023)Zheng, Chiang, Sheng, Zhuang, Wu, Zhuang, Lin, Li, Li, Xing, Zhang, Gonzalez, and Stoica}]{zheng-2023-judging}
Lianmin Zheng, Wei-Lin Chiang, Ying Sheng, Siyuan Zhuang, Zhanghao Wu, Yonghao Zhuang, Zi~Lin, Zhuohan Li, Dacheng Li, Eric Xing, Hao Zhang, Joseph~E. Gonzalez, and Ion Stoica. 2023.
\newblock \href {https://openreview.net/forum?id=uccHPGDlao} {Judging {LLM}-as-a-judge with {MT}-bench and chatbot arena}.
\newblock In \emph{Thirty-seventh Conference on Neural Information Processing Systems Datasets and Benchmarks Track}.

\end{thebibliography}
